\documentclass{article}
\usepackage{spconf,amsmath,graphicx}
\usepackage{amssymb}
\usepackage{todonotes}
\usepackage{tabulary}
\usepackage{tabularx}
\usepackage{multirow}
\usepackage{blindtext}
\usepackage{url}

\usepackage{stfloats}

\def\x{{\mathbf x}}
\def\y{{\mathbf y}}
\def\b{{\mathbf b}}

\title{Implicit Modeling with Uncertainty Estimation\\ for Intravoxel Incoherent Motion Imaging}
\name{Lin Zhang$^1$, Valery Vishnevskiy$^1$, Andras Jakab*$^2$, Orcun Goksel*$^1$\thanks{*These authors contributed equally}}
\address{ $^1$Computer-assisted Applications in Medicine, ETH Zurich, Switzerland\\
    $^2$Center for MR-Research, University Children's Hospital Zurich, Switzerland}

\begin{document}
\maketitle
\begin{abstract}
Intravoxel incoherent motion (IVIM) imaging allows contrast-agent free in vivo perfusion quantification with magnetic resonance imaging (MRI). 
However, its use is limited by typically low accuracy due to low signal-to-noise ratio (SNR) at large gradient encoding magnitudes as well as dephasing artefacts caused by subject motion, which is particularly challenging in fetal MRI.
To mitigate this problem, we propose an implicit IVIM signal acquisition model with which we learn full posterior distribution of perfusion parameters using artificial neural networks. 
This posterior then encapsulates the uncertainty of the inferred parameter estimates, which we validate herein via numerical experiments with rejection-based Bayesian sampling.
Compared to state-of-the-art IVIM estimation method of segmented least-squares fitting, our proposed approach improves parameter estimation accuracy by 65\% on synthetic anisotropic perfusion data.
On paired rescans of in vivo fetal MRI, our method increases repeatability of parameter estimation in placenta by 46\%.

\end{abstract}

\section{Introduction}
\label{sec:intro}
Diffusion-weighted imaging (DWI) is a non-invasive method to generate MRI contrast based on water molecule diffusion~\cite{merboldt1991diffusion}. 
In the classic DWI, signal attenuation is explained by a mono-exponential decay. 
However, multiple micro-scale translational motions may contribute to DWI signals, such as microcirculation (perfusion) of blood.
IVIM imaging aims to disentangle the diffusion and perfusion of water molecules, by assuming a bi-exponential decay model~\cite{le1988separation}. 

In recent years, there has been increasing interest in adapting this technique to prenatal diagnostic imaging. 
Previous studies revealed that assessing the placenta perfusion from IVIM imaging may provide valuable diagnostic information about fetal growth~\cite{alison2013use}. 
In~\cite{jakab2017intra,jakab2018microvascular} IVIM imaging of other developing organs such as fetal liver, lungs, kidneys and brain has been investigated. 
These studies indicate a potential link between IVIM parameters and gestational changes in microstructural development.

Successful clinical translation of prenatal IVIM imaging is severely limited by poor parameter estimation accuracy due to low SNR of clinical fetal MR images and unpredictable fetal movements in utero.
Estimating IVIM parameters from observed signal is a challenging \emph{inverse mapping} problem.
Several algorithms have been proposed to increase estimation quality, such as maximum likelihood estimation (MLE)~\cite{notohamiprodjo2015combined} and Bayesian fitting~\cite{spinner2017bayesian}. 
Due to likelihood nonlinearities, the MLE accuracy is limited by low SNR~\cite{zhang2013cramer} of DWI.
Although the Bayesian fitting approach~\cite{spinner2017bayesian} can be more robust to noise, it requires the manual delineation of regions that are assumed to be homogeneous, aiming only for quantification. 

Recently, neural networks have achieved great success in many tasks such as classification and regression. 
An artificial neural network approach was proposed in~\cite{bertleff2017diffusion} for IVIM inverse parameter mapping, which is trained to provide robust and fast maximum likelihood estimates.
In practice, this approach may achieve higher accuracy and tissue contrast compared to conventional least-squares fitting, but is affected by the same theoretical limitations of MLE.
Moreover, a point estimate cannot provide sufficient information about an inverse mapping, particularly for stochastic models.

Inverse mapping uncertainty quantification was studied in~\cite{zhang2013cramer} by analyzing the variance of MLE using Cram{\'e}r-Rao lower bound. 
However, for nonlinear parameter estimation the estimator variance does not necessarily reflect the true uncertainty of inverse mapping caused by the stochastic nature of signal acquisition. 
To address this, we herein propose to model inverse-mapping uncertainty from the inference of entire posterior distribution using Gaussian approximation.
A neural network is trained to stochastically maximize the log-likelihood of the input-dependent mean and variance of this Gaussian distribution. 
We show below that such verbose estimation of the posterior allows to better reflect uncertainty in the data acquisition process and helps to improve repeatability of clinical fetal IVIM estimation.

\section{Background}
\label{sec:theoryandmethod}
IVIM model explains signal attenuation as a weighted sum of diffusion $D$ and intravoxel perfusion (pseudo-diffusion) $D^\ast$, weighted by perfusion fraction $f$.
Trace signal $S$ for b-value $b_i$ is then proportional to the unweighted signal $S_0$ as follows:
\begin{align}
\label{eq:ivim_tr}
    S(b; \y)=S_0\left(f\exp(-bD^\ast)+(1-f)\exp(-bD)\right)\,,
\end{align}
leading to $n_p=4$ model parameters to estimate $\y=\{S_0,f,D,D^\ast\}\in\mathbb{R}^{n_p}$.
Given observed signal $\x$$\in$$\mathbb{R}^{n_b}$ for $n_b$ b-values $\mathbf{b}$, model~(\ref{eq:ivim_tr}) can be fitted in least-squares sense:
 \begin{align}
     \hat{\y}_\text{LS}=\text{argmin}_\y \|S(\b; \y)-\x\|_2^2,
 \end{align}
which is the MLE with Gaussian noise assumption.
Due to multiple local minima and high nonlinearity, a two-stage least-squares fitting (LSQ) procedure was applied to increase stability~\cite{notohamiprodjo2015combined}.
In the first stage, a mono-exponential diffusion model is fitted at high b-values 
($b$$\geq$$250$s$/\text{mm}^2$), 
for which perfusion effects are assumed to be negligible.
This fixes diffusion $D$ and $S_0$ estimates.
Given these, in the second step, the entire model is fitted for all b-values to estimate perfusion $D^\ast$ and its fraction $f$.
In practice, box constraints are used to avoid infeasible parameter values~\cite{spinner2017bayesian}. 
Accurate estimation with this approach still requires high SNR~\cite{zhang2013cramer}.

\section{Methods}
Normality of signal magnitude is valid only for high SNR at low b-values. 
Explicit likelihood with Rician noise model, signal averaging and dephasing would thus be intractable using MLE.
However, such distributions can be constructed \emph{implicitly} by defining the sampling procedure for $p(\x|\y)$ as:
\begin{align}
\label{eq:p_x_sample}
    x_i\leftarrow
    \frac{1}{n_g}\sum_{j\leq n_g}\text{Rice}(S(b_i; \mathbf{y})\alpha^{\gamma_i}, \nu_j), \; 
    \alpha\sim\mathcal{U}(0, 1),\;
\end{align}
where $\gamma_i\in\{0,1\}$ is Bernoulli distributed indicator that gates uniformly distributed attenuation factor $\alpha$. 
This ad~hoc model simulates signal dephasing due to object motion, that can affect signal with probability $\beta_i$, depending on the b-value~\cite{stoeck2018ismrm}.
Signal component for each of $n_g$ diffusion gradients is assumed to be Rician with noise level $\nu_j$.
To construct the sampling process for paired data $p(\x, \y)=p(\x|\y)p(\y)$, we define prior distribution $p(\y)$ to be uniform over the physically admissible set of model parameters, similarly to box constrains employed in LSQ~\cite{spinner2017bayesian}.

The posterior over model parameters $p(\y|\x)$ is the object of interest and is formally defined by $p(\x,\y)$. 
Although approximate rejection algorithms that allow to sample from posterior exist~\cite{csillery2010approximate}, this process requires exponential (w.r.t. $n_p$) number of simulations per observed signal $\x$, and thus is infeasible for imaging in practice.

\noindent
\textbf{3.1. Amortized Gaussian posterior.}
Following~\cite{nix1994estimating}, posterior is modelled to be Gaussian with mean and diagonal covariance matrix parametrized by signal-dependent  multi-layer perceptrons (MLP) $\boldsymbol{\mu}$ and $\boldsymbol{\lambda}$ with tunable weights $\Theta$:
\begin{align}
\label{eq:gauss_posterior}
    p_\Theta(\mathbf{y}|\mathbf{x}) \triangleq 
    \mathcal{N}\left(
        \boldsymbol{\mu}(\mathbf{x};\Theta),
        \text{diag}\left( \exp\left( \boldsymbol{\lambda}\left( \x;\Theta \right)\right)\right)
    \right)\,,
\end{align}
where we perform exponential reparametrization of the covariance to guarantee non-negativity, while $\boldsymbol{\mu}$ and $\boldsymbol{\lambda}$ map signal to parameter space: $\mathbb{R}^{n_b}\rightarrow\mathbb{R}^{n_g}$.

\begin{figure}[t]
\centering
{\includegraphics[width=.8\columnwidth]{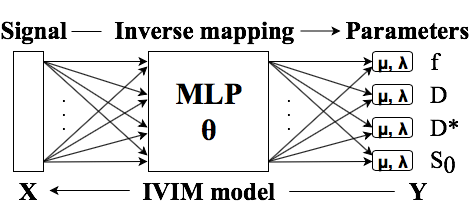}
}
\caption{AGP network architecture for inverse IVIM mapping.
} 
\label{fig:network}
\end{figure}

Minimizing expected KL-divergence between true and variational posteriors
$\mathop{\mathbb{E}}_{p(\x)} \mathcal{D}_\text{KL}(p(\y|\x)||p_{\Theta}(\y|\x))$
is equivalent to maximizing the log-likelihood of predictions:
\begin{align}
    \label{eq:GLL}
    \mathcal{L}(\Theta)\!\!=\!\!\!\!{\displaystyle \mathop{\mathbb{E}}_{p(\mathbf{x},\mathbf{y})}} \sum_{j\leq n_p} \!\!\!-\lambda_j(\mathbf{x};\Theta) - 
    \frac{(y_j-\mu_j(\mathbf{x};\Theta))^2}{2\exp(2\lambda_j(\mathbf{x};\Theta))},
\end{align}
which can be achieved using stochastic optimization algorithms, given the efficient sampling procedure defined in~(\ref{eq:p_x_sample}).
Our network architecture is shown in Fig~\ref{fig:network}. 
We use a fully connected MLP with 5 hidden layers and 50 nodes per layer. 
Hyperbolic tangent activation is used at each layer except the output layer, which is linear. 

Our proposed IVIM inverse mapping, hereafter called Amortized Gaussian Posterior (AGP), uses the model $p_\Theta(\y|\x)$ in~(\ref{eq:gauss_posterior}) that is trained via optimization of~(\ref{eq:GLL}).

\section{Experiments and Results}
\noindent\textbf{4.1. Network training.}
For network training, signals were simulated according to~(\ref{eq:p_x_sample}) also modeling the intended use, acquisition and sequence settings.
We assumed uniform prior distribution $p(\mathbf{y})$ with the following parameter ranges:
$S_0$$\in$ $[0,3000]$, 
$f$$\in$$[0.0005,0.9995]$, 
$D$$\in$$[0.045, 5]$$\times$$10^{-3}$  $\text{mm}^2/\text{s}$, and $D^\ast$$\in$$[0.34, 100]$$\times$$10^{-3}$   $\text{mm}^2/\text{s}$, 
ensuring that $D$$\leq$$D^\ast$.
To simulate the in vivo acquisition setup described later below, we use Rician noise with $\nu\in[6,18]$ depending on the diffusion encoding gradient.
Different noise levels originate from the image reconstruction procedure of the scanner and the use of tetrahedral gradients. 
Dephasing probability $\gamma_i$ was chosen according to the expected in vivo data quality and is b-value dependent: $2\%$ for $b$$<$$300$\,$\text{s}/\text{mm}^2$ and linearly increased from $10\%$ to $25\%$ for $b$$\geq$$300$\,$\text{s}/\text{mm}^2$.
The averaged (trace) signals, generated on-the-fly during training, were used as input to the network. 
We trained the network for $10^6$ iterations using the Adam optimizer~\cite{kingma2014adam}, with a learning rate of $10^{-3}$ and a batch size of $2000$ samples.

\begin{figure*}
{ 
  {\includegraphics[width=4.0cm]{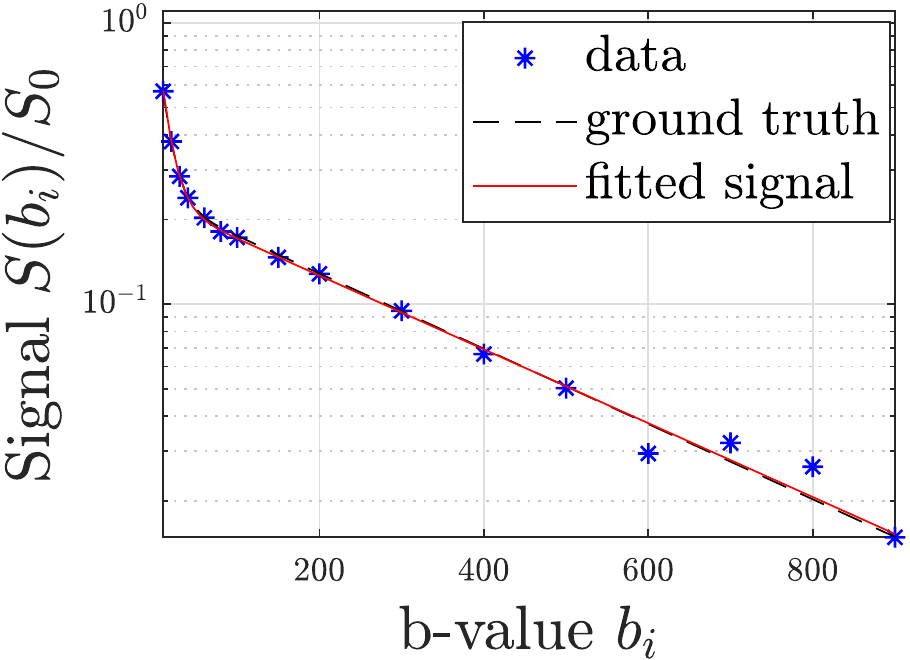}} \hspace{3mm}
  {\includegraphics[width=4.0cm]{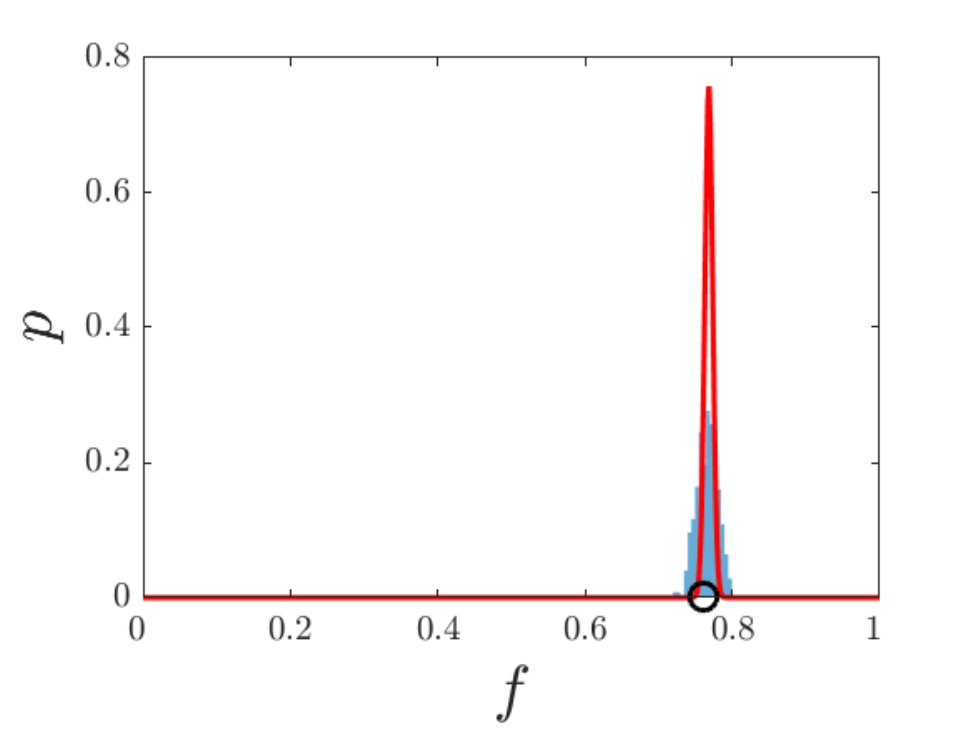}} \hspace{3mm}
  {\includegraphics[width=4.0cm]{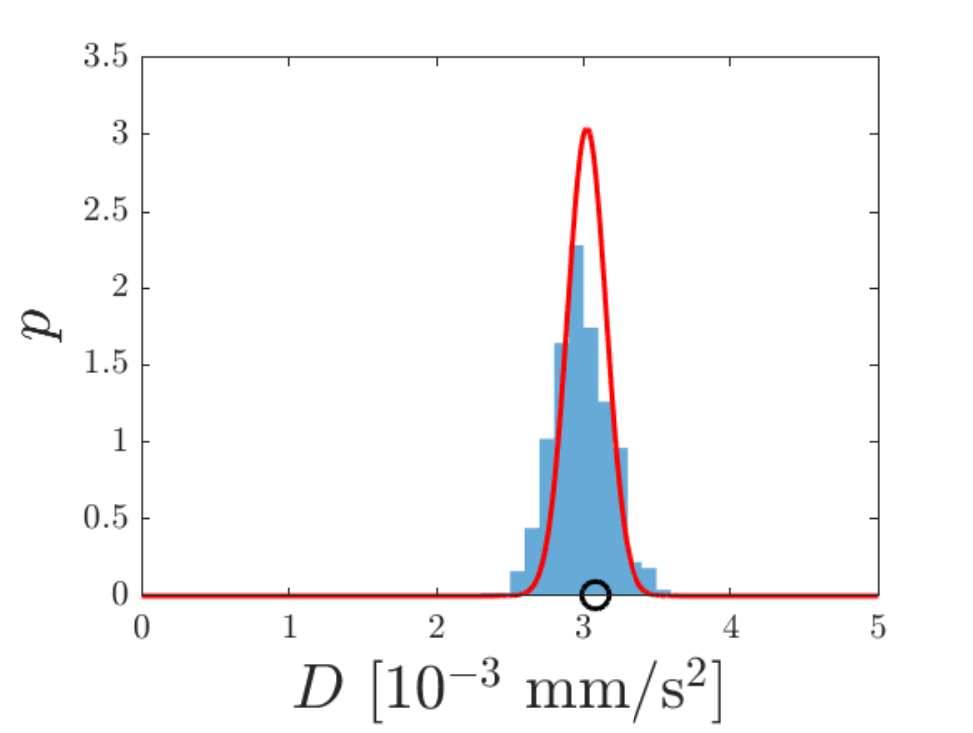}} \hspace{3mm}
  {\includegraphics[width=4.0cm]{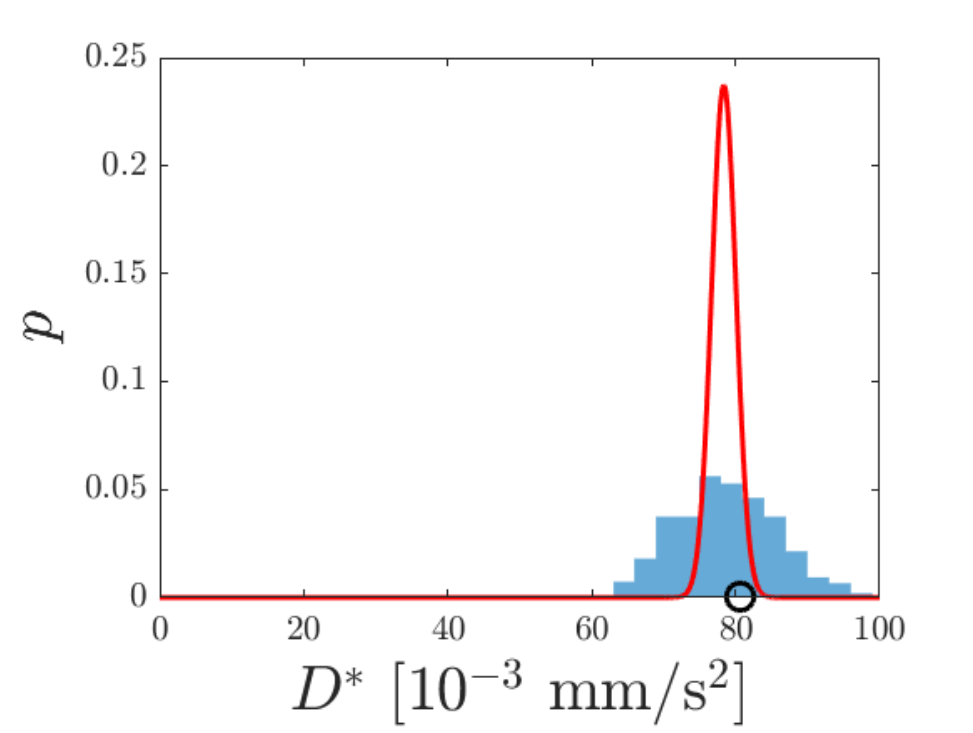}}}\\
  {\includegraphics[width=4.0cm]{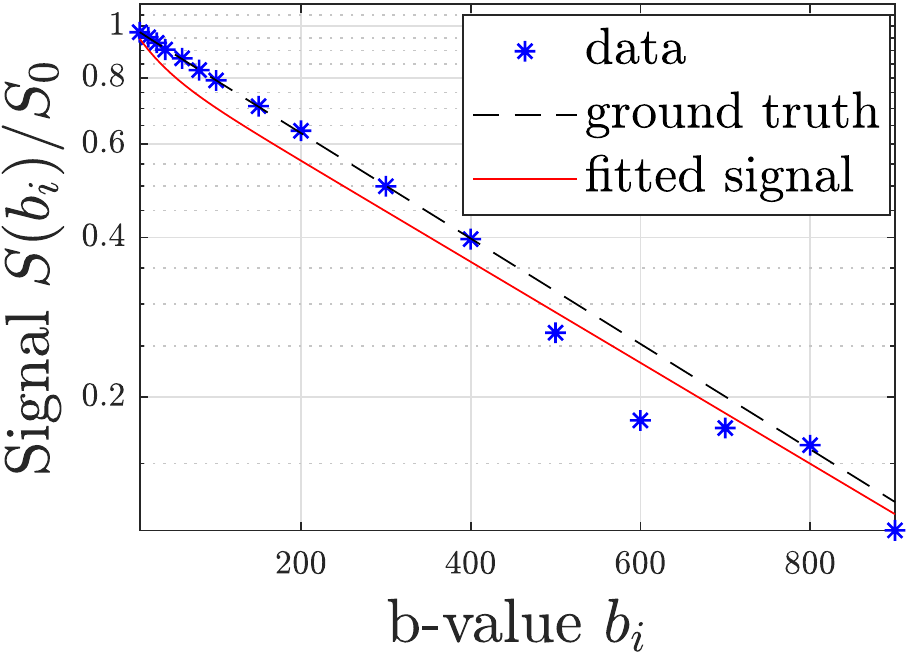}} \hspace{3mm}
  {\includegraphics[width=4.0cm]{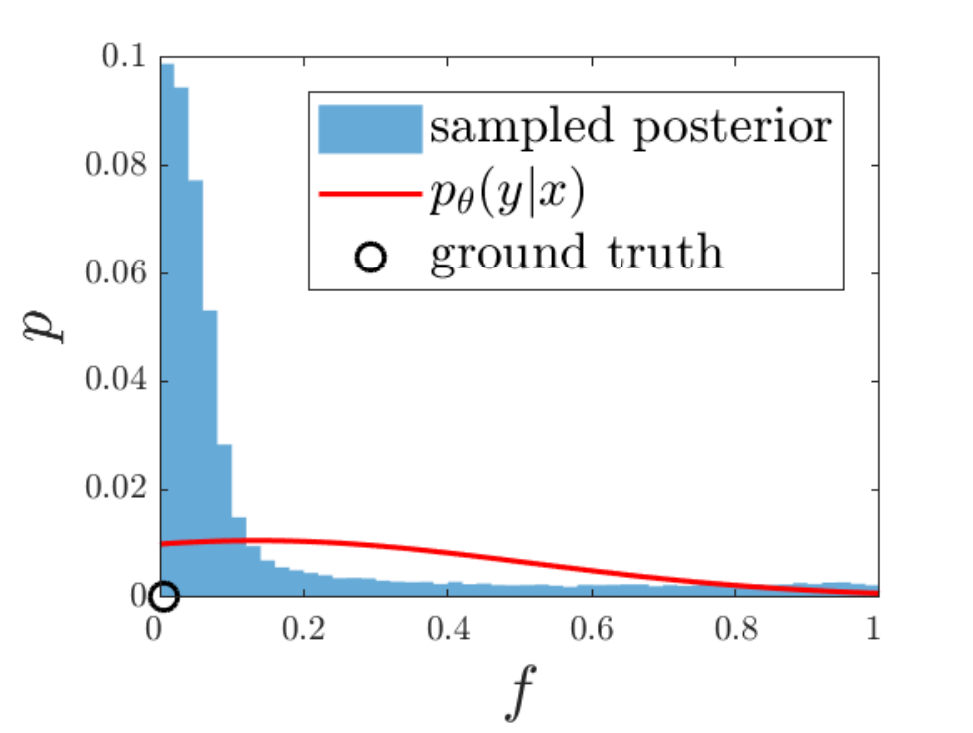}} \hspace{3mm}
  {\includegraphics[width=4.0cm]{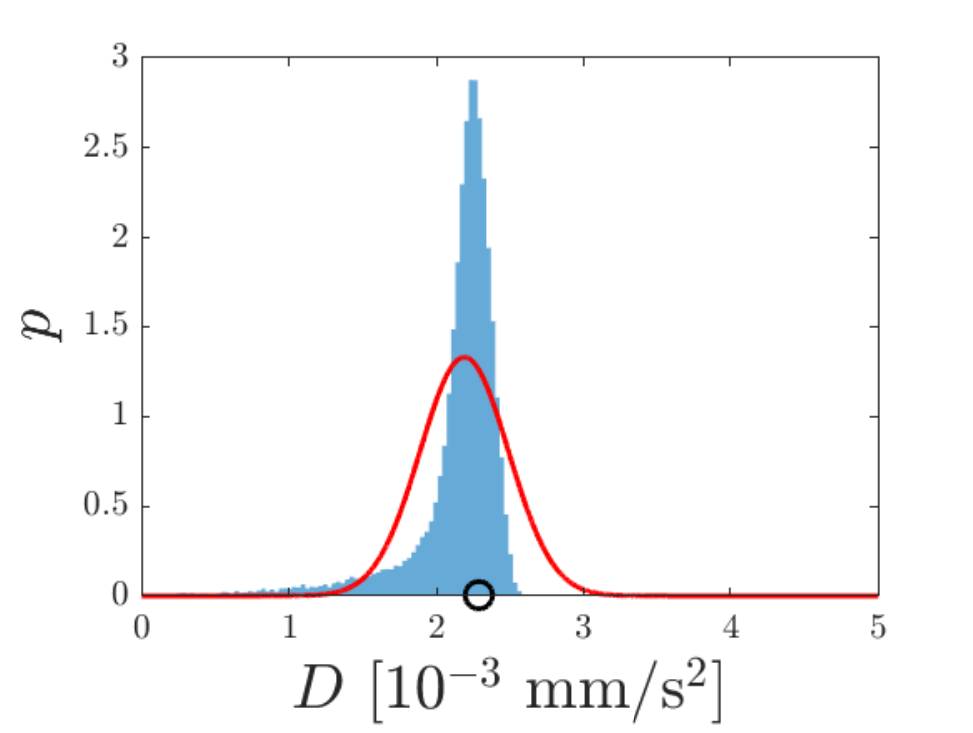}} \hspace{3mm}
  {\includegraphics[width=4.0cm]{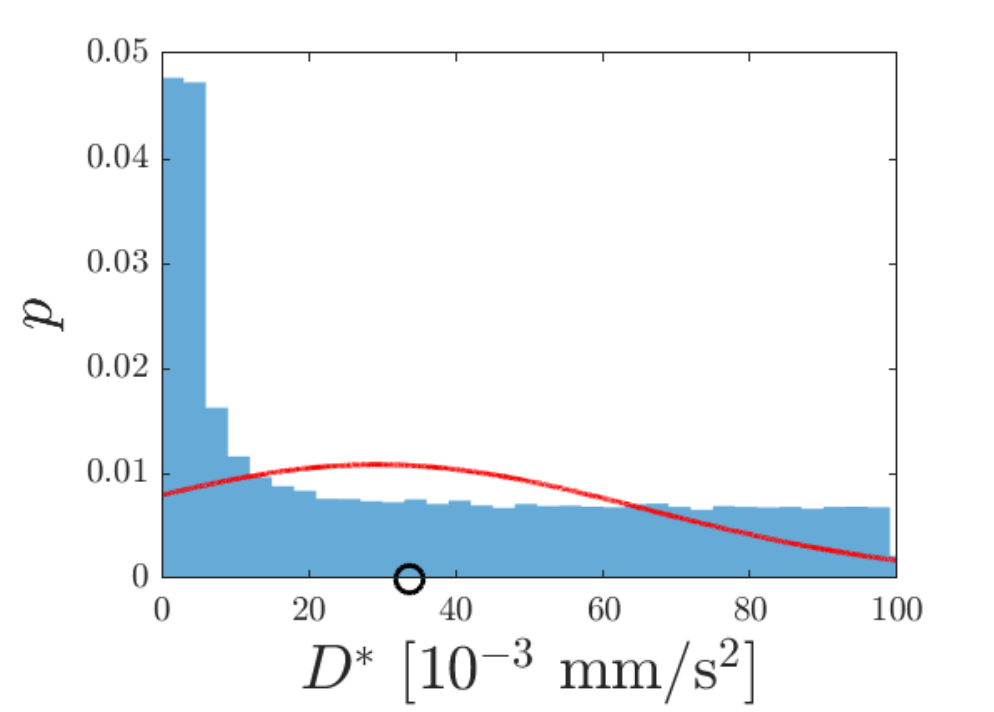}}
  \caption{Two simulated IVIM signals, showing the agreement of their sampled~\cite{csillery2010approximate} and AGP-predicted posteriors.  %
  } 
  \label{fig:posterior_plots}
\end{figure*}

\noindent\textbf{4.2. IVIM uncertainty quantification via AGP.}
In this experiment we verify the inverse mapping inferred by AGP on the synthetic data generated by procedure~(\ref{eq:p_x_sample}). 
We report the estimation uncertainty as the corresponding standard deviation of the predictive posterior defined in~(\ref{eq:gauss_posterior}).
Two sample scenarios are seen in Fig.~\ref{fig:posterior_plots}.
The top row illustrates a configuration where perfusion and diffusion effects can be disentangled with relative certainty.
The bottom row demonstrates a case for low certainty in estimating $D^\ast$, which is likely caused by the low perfusion fraction $f$, i.e. the perfusion contribution to the signal being overwhelmed by acquisition noise, hence no reliable estimation being possible.

\begin{figure}[t]
{\includegraphics[width=2.725cm]{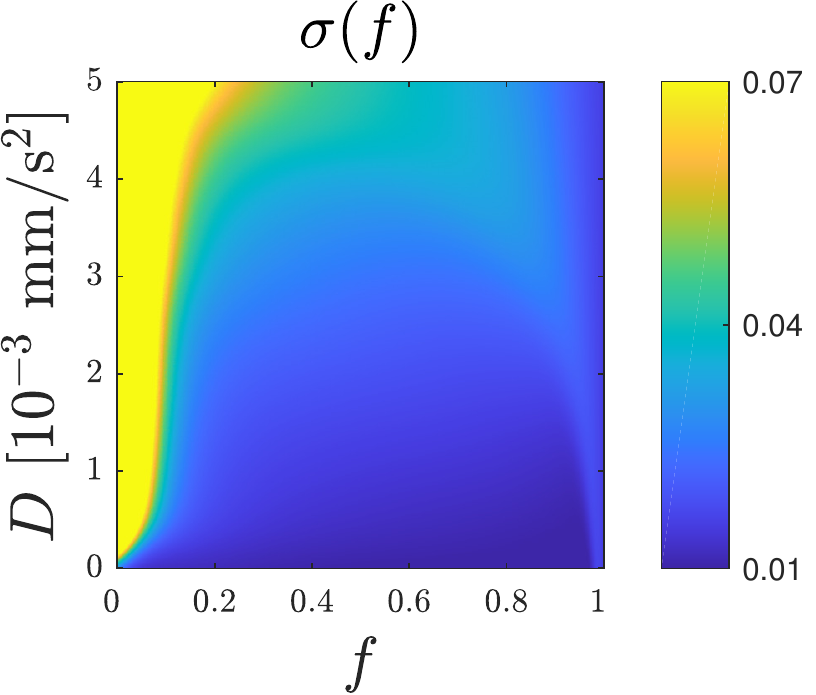}\hspace{2mm}
\includegraphics[width=2.6cm]{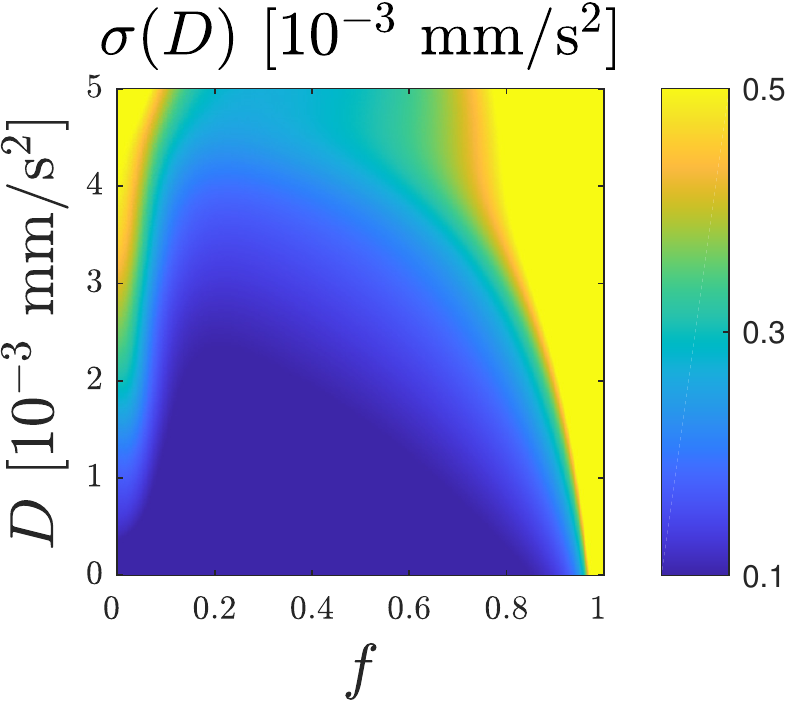}\hspace{2mm}
\includegraphics[width=2.6cm]{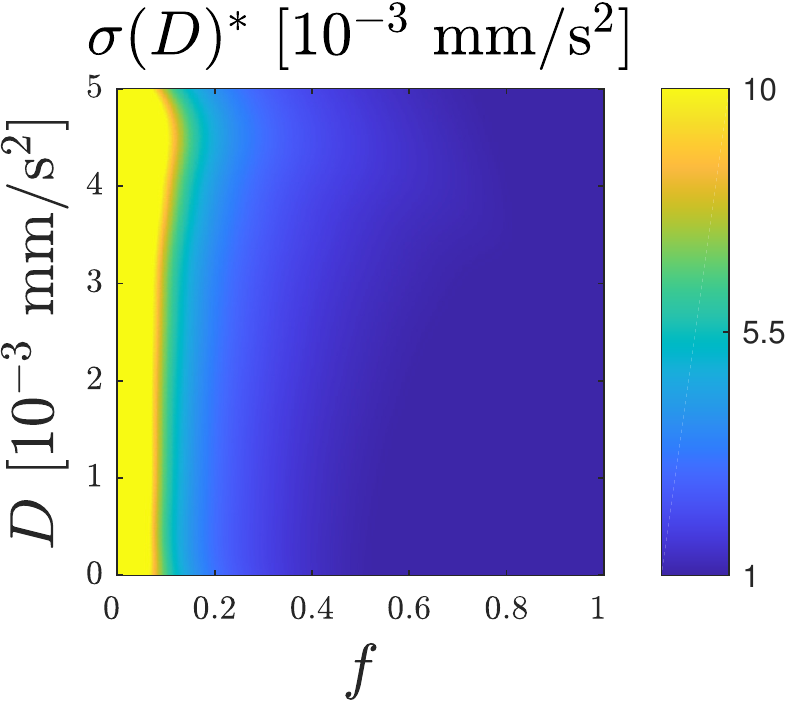}\\
\includegraphics[width=2.725cm]{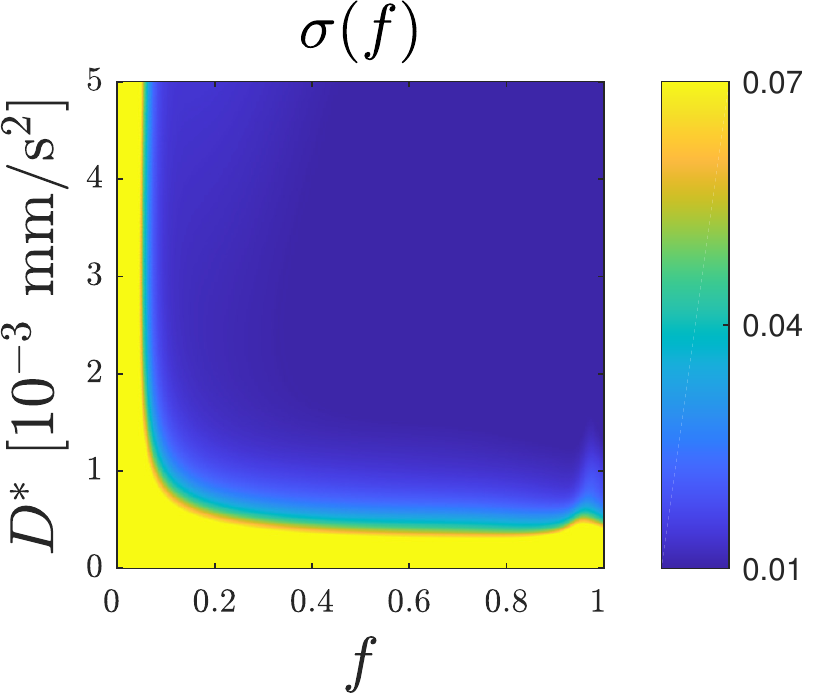}\hspace{2mm}
\includegraphics[width=2.6cm]{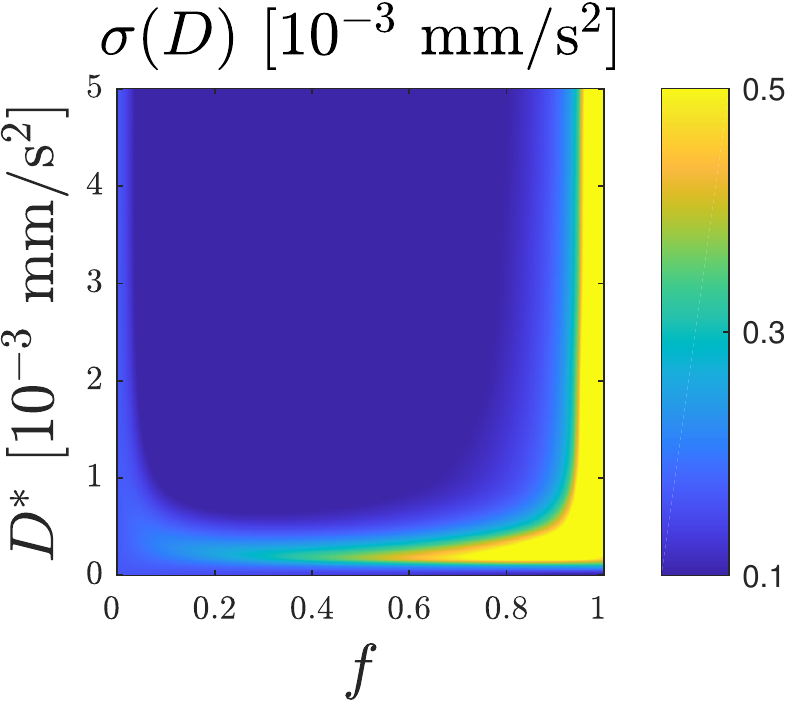}\hspace{2mm}
\includegraphics[width=2.6cm]{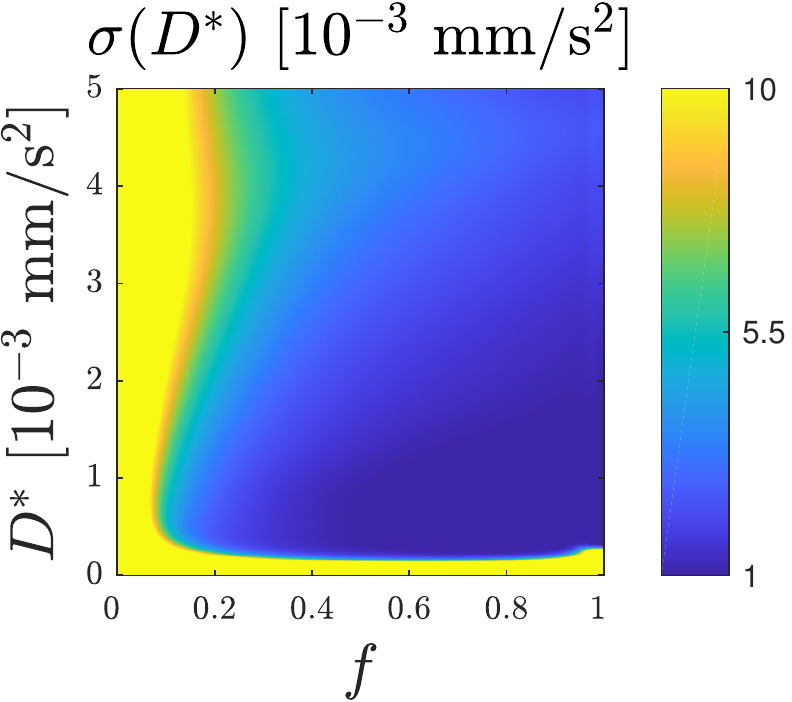}\\
\includegraphics[width=2.725cm]{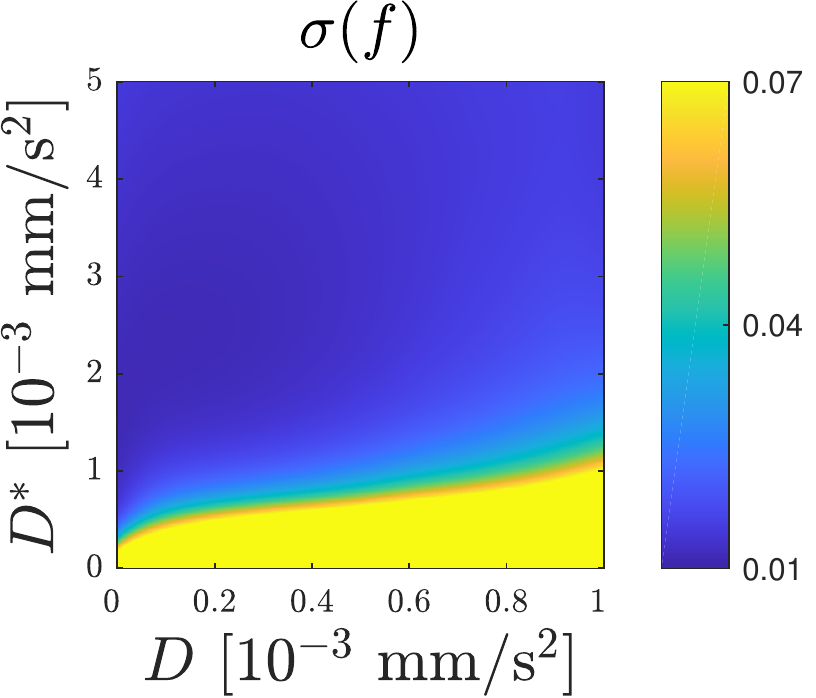}\hspace{2mm}
\includegraphics[width=2.6cm]{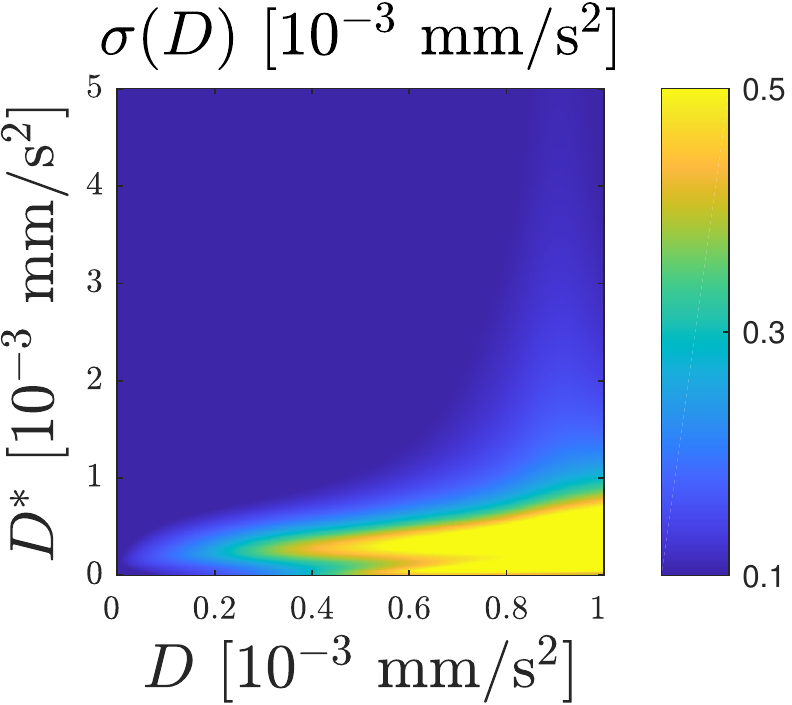}\hspace{2mm}
\includegraphics[width=2.6cm]{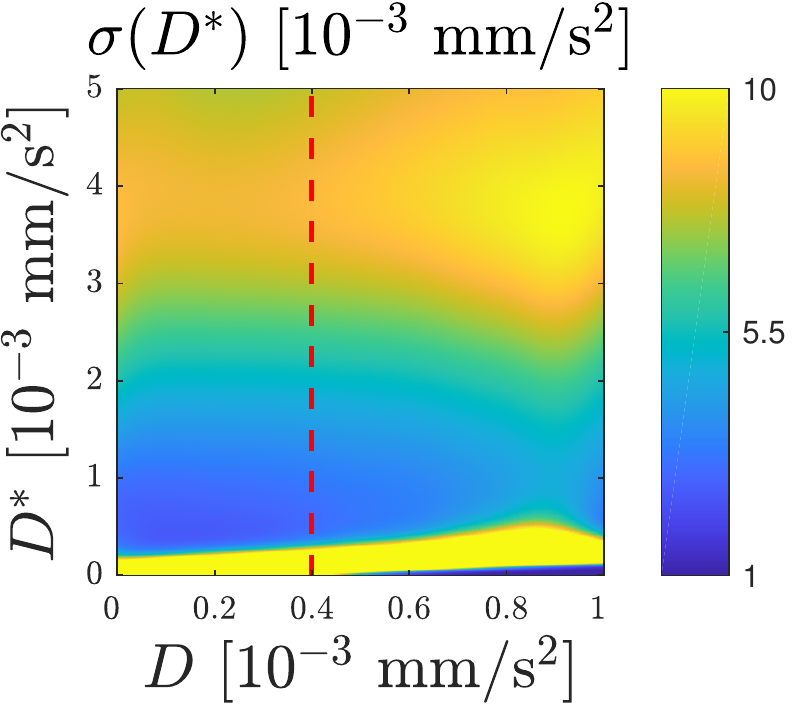}
}
\caption{{Estimation uncertainty with different parameter combinations inferred by AGP.} $\sigma$ is the standard deviation of the predictive posterior. Top row: 
$D^\ast$$=$$0.02$ $\text{mm}^2/\text{s}$, middle: 
$D$$=$$0.001$ $\text{mm}^2/\text{s}$, bottom: 
$f$$=$$0.2$.
$S_0$=$1000$.} 
\label{fig:heat_map}
\end{figure}

Inverse mapping uncertainty of IVIM parameters $f$, $D$ and $D^\ast$ with varying simulated parameter combinations are depicted in
Fig.~\ref{fig:heat_map}.
In each experiment, one parameter was fixed while the other two were varied in 1000 steps within their ranges. 
Each estimation uncertainty is the averaged standard deviation of predictive posteriors over 1000 different noise realizations.
In results, high uncertainty is observed when $D$ and $D^\ast$ values are similar or when $f$ approaches one or zero. 
The estimation uncertainty behaviour for $D^\ast$ is well illustrated along the red dotted line in the bottom-right plot in Fig.~\ref{fig:heat_map}: The uncertainty is seen to increase with larger values of $D^\ast$, since the perfusion contribution to the signal $\exp(-bD^\ast)$ approaches zero; while, again a high uncertainty is observed for smaller values of $D^\ast$, but this time because of perfusion and diffusion effects becoming indistinguishable.

\noindent\textbf{4.3. Simulated anisotropic data.}
Original isotropic IVIM model~\cite{le1988separation} considered in~(\ref{eq:ivim_tr}) relies on the fact that trace of the signal approximates directionally averaged contributions of diffusion and perfusion, which can be more accurately modeled as tensors~\cite{finkenstaedt2017ivim}.
By simulating anisotropic IVIM data for testing of models trained using only isotropic data~(\ref{eq:p_x_sample}), we aim to demonstrate the sensitivity of AGP and LSQ to modelling assumptions.
We simulate 1024 randomly rotated diffusion tensors with 
$\text{tr}(\mathbf{D})$$=$$9.4$$\cdot$$10^{-4}$$\text{mm}^2/\text{s}$, 
$\text{tr}(\mathbf{D^\ast})$$=$$5.3$$\cdot$$10^{-2}$$\text{mm}^2/\text{s}$, fractional anisotropy of 0.8 and $f$$=$$0.18$. 
The observed signal is simulated using tetrahedral diffusion gradient configuration~\cite{conturo1996encoding}, followed by dephasing and Rician noise injection similarly to (\ref{eq:p_x_sample}).
It is assumed that inverse mapping would recover rotation invariant traces of the tensors that are taken as the ground truth~($D$$\approx$$\text{tr}(\mathbf{D})$).

We compare our method AGP with the segmented LSQ approach. 
Mean absolute errors between the estimated and true parameters are reported in Table~\ref{tbl:aniso_res}, showing that AGP achieves smaller estimation errors in $D^\ast$ and comparable results in $f$ and $D$ compared to LSQ for the signals modeled without dephasing. 
For signals with dephasing, AGP outperforms the LSQ approach substantially.
\begin{table}[b]
		\begin{center}
			\caption{Mean absolute error (MAE) between ground truth and estimated IVIM parameter values by AGP and LSQ on the simulated anisotropic data with out (w/o) and with (w) dephasing (dp). Bold numbers indicate lower MAE.
			}
			\label{tbl:aniso_res}
			\begin{tabulary}{\linewidth}{l|c|l|c|c}
				\hline 
				\multicolumn{3}{c}{~} & \textbf{LSQ} & \textbf{AGP}\\
				\hline
				\multirow{6}{*}{\rotatebox{90}{\textbf{MAE}}} 
				& \multirow{2}{*}{$f\,[\%]$} & w/o dp & 2.6 & \textbf{2.2}\\
				                         &  & w. dp & 8.4 & \textbf{2.0}\\\cline{2-5}
				& \multirow{2}{*}{$D$\,[$10^{-4}$\,mm/s$^2$]} & w/o dp & \textbf{1.69} & 1.81\\
				                              &  & w. dp & 2.27 & \textbf{1.30}\\\cline{2-5}
				&\multirow{2}{*}{$D^\ast$\,[$10^{-3}$ mm/s$^2$]} & w/o dp & 29.3 & \textbf{11.5}\\
				                                    &  & w. dp & 40.1 & \textbf{9.8}\\\hline
			\end{tabulary}
		\end{center}
	\end{table}

\noindent
\textbf{4.4. In vivo fetal MRI. }
Prenatal MRI was performed on a 1.5\,T Discovery MR450 unit (GE Healthcare, Milwaukee, WI, USA) on 17 subjects. 
The images were acquired with a dual spin-echo planar sequence, echo time of $2200/75$\,ms, acquisition matrix $80$$\times$$100$, voxel size of $2$$\times$$2$\,mm$^2$, slice thickness $3$ or $4$\,mm.
The number of slices varies from 9 to 24.
Tetrahedral diffusion gradients~\cite{conturo1996encoding} were used with 16 b-values: 10, 20, 30, 40, 60, 80, 100, 150, 200, 300, 400, 500, 600, 700, 800, 900 $\text{mm}/\text{s}^2$. 
One $b_0$ image was acquired.
IVIM imaging sequence was repeated twice for each subject. 

Prior to model fitting, we applied a state-of-the-art registration method~\cite{vishnevskiy2017isotropic} using a cost function based on principle component analysis (PCA)~\cite{huizinga2016pca} to perform group-wise registration. %
First, images with the same b-value but different diffusion gradients were aligned and averaged slice-wise to construct a single image volume with high SNR. 
Then, a 3D group-wise registration was applied to align these averaged volumes of 16 different b-values.

\begin{figure}[t]
\centering
\includegraphics[width=2.6cm]{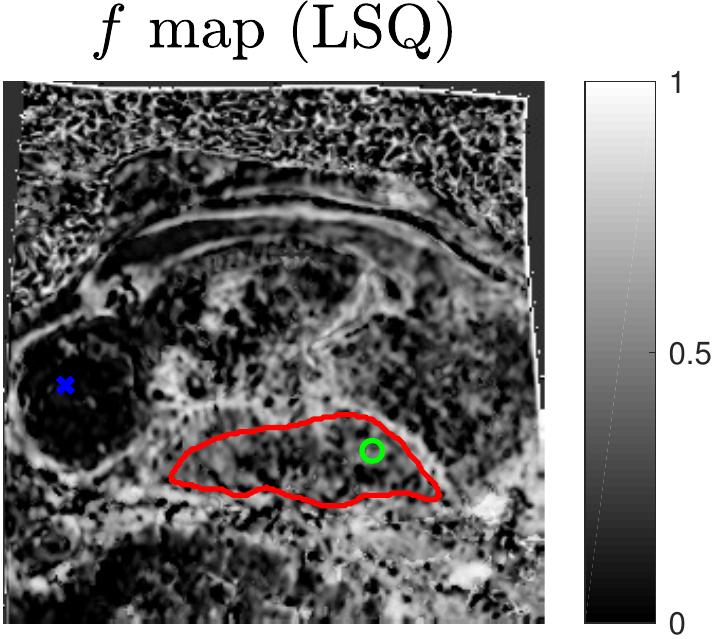}\hfill
\includegraphics[width=2.82cm]{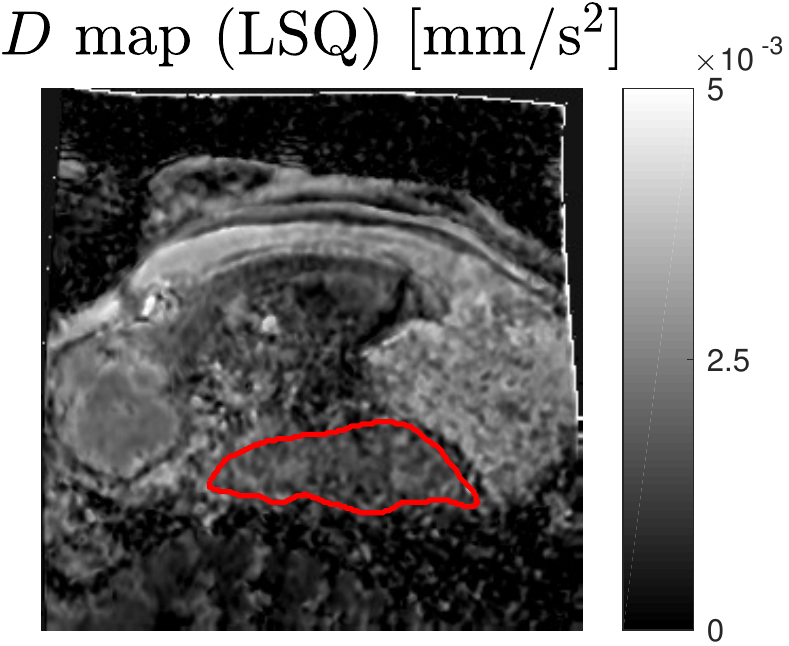}\hfill
\includegraphics[width=2.80cm]{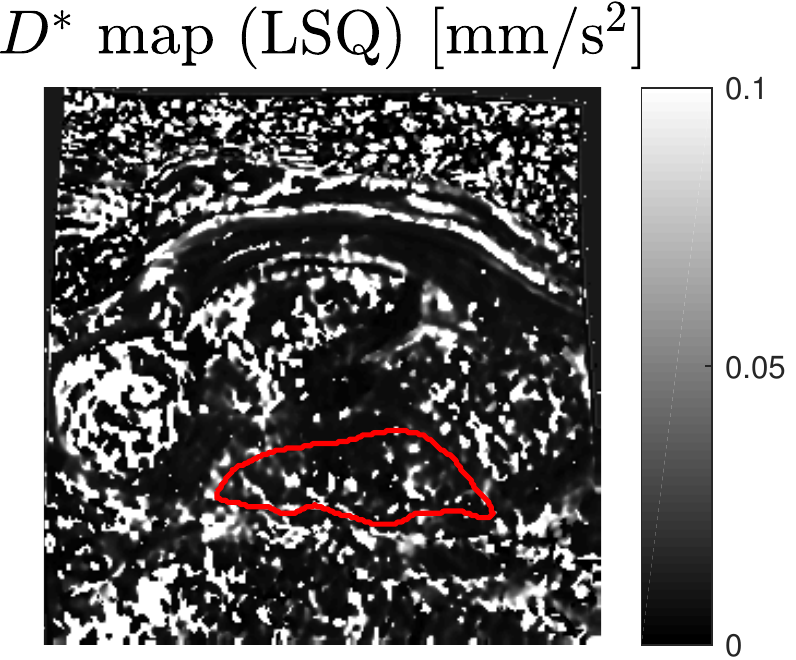}\\
\includegraphics[width=2.6cm]{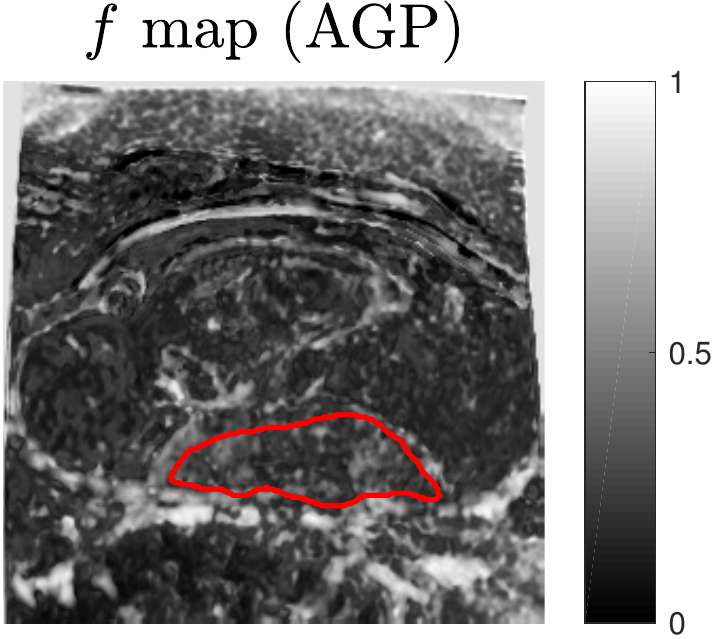}\hfill
\includegraphics[width=2.83cm]{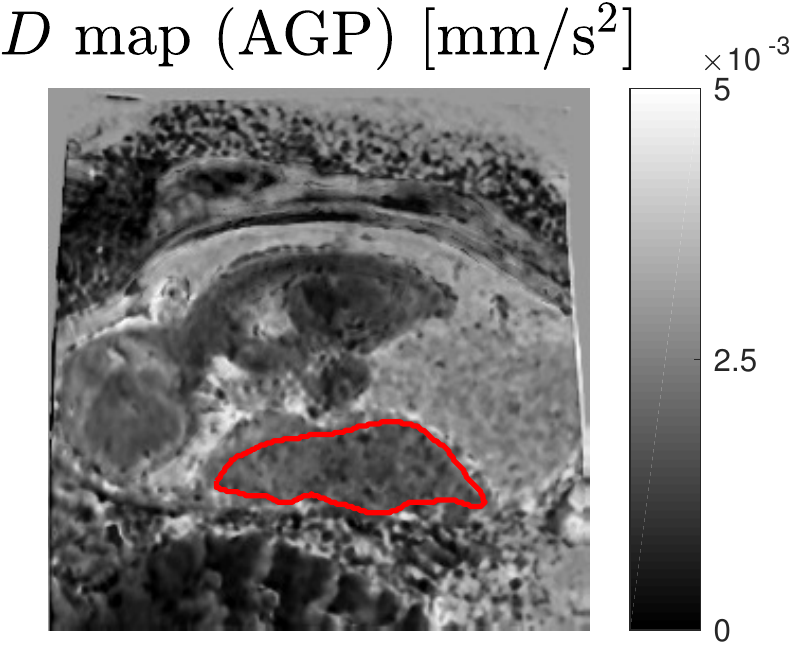}\hfill
\includegraphics[width=2.85cm]{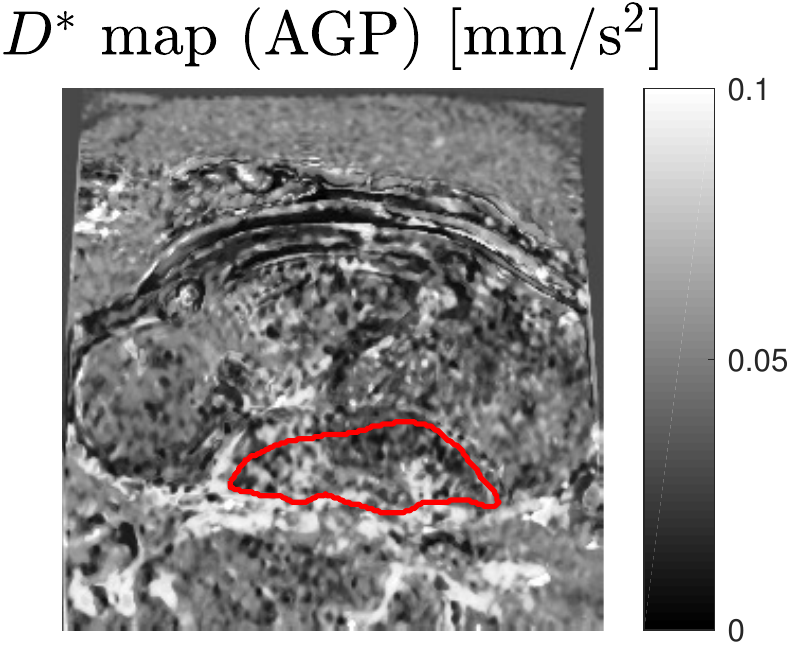}\\
\includegraphics[width=2.58cm]{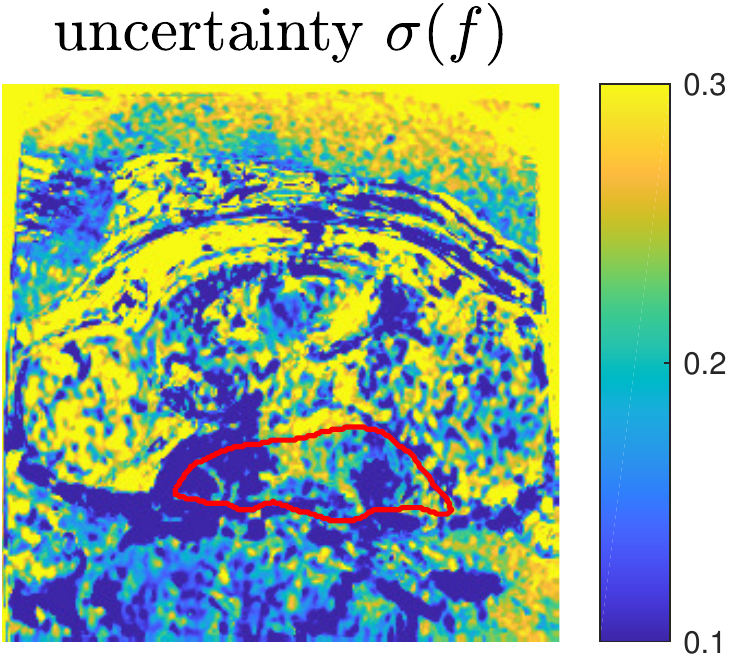}\hfill
\includegraphics[width=2.95cm]{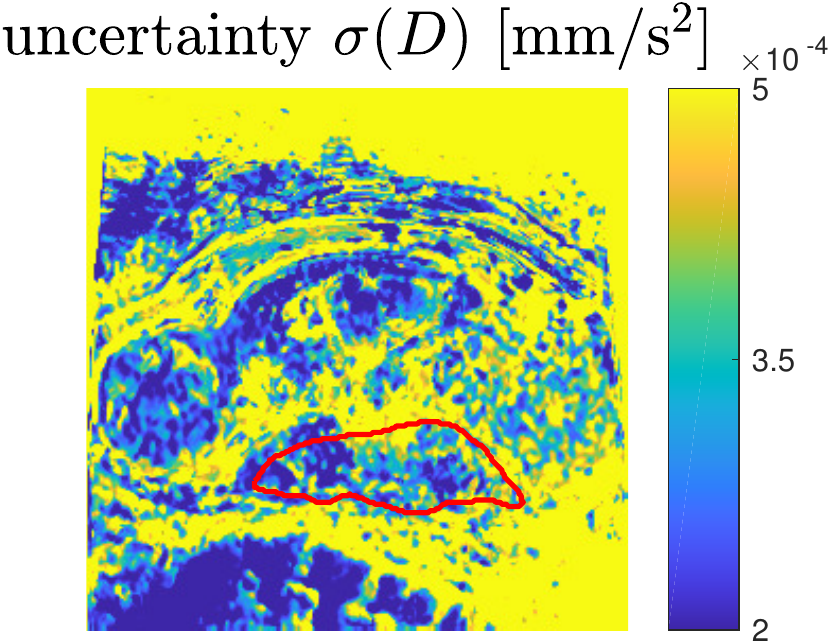}\hfill
\includegraphics[width=2.95cm]{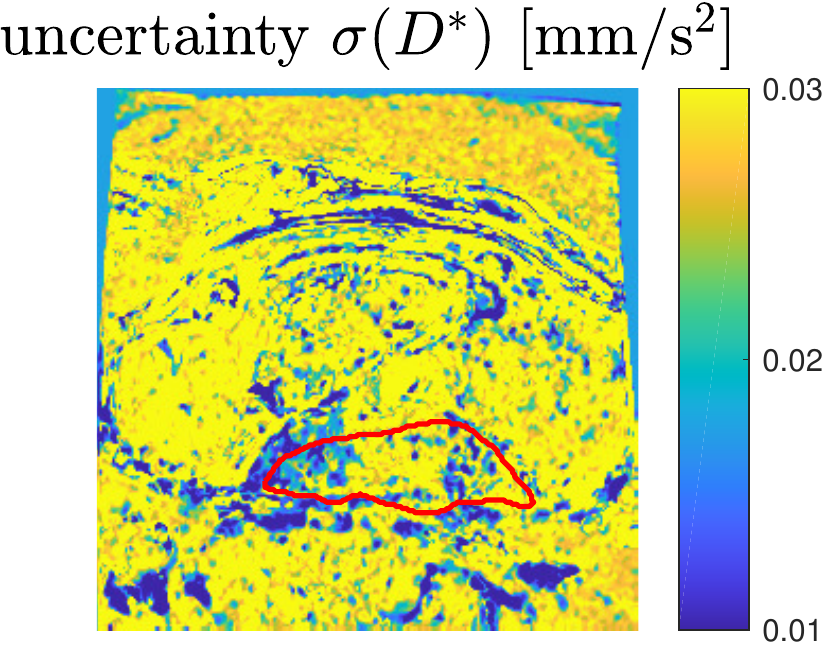}\\[1ex]
\includegraphics[width=4cm]{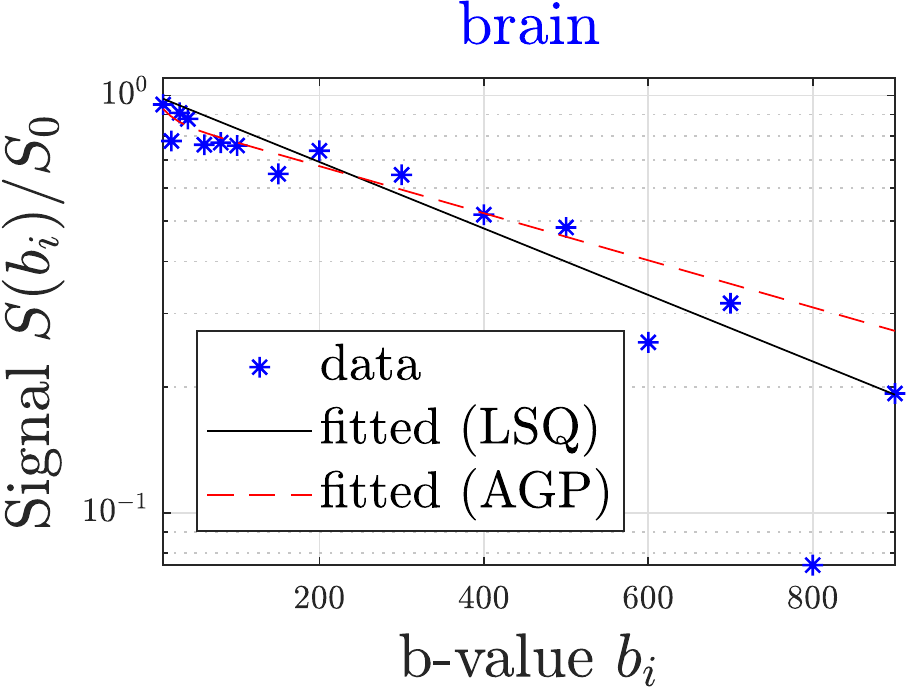}\hfill
\includegraphics[width=4cm]{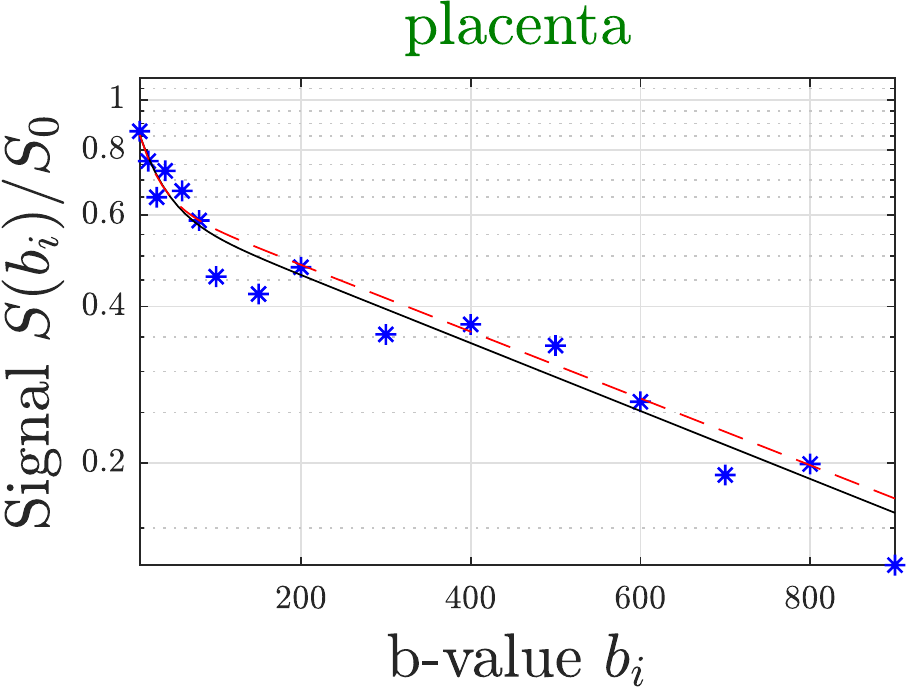}
\caption{{An in vivo fetal IVIM case, showing LSQ and AGP estimates, and AGP uncertainties. Placenta is outlined in red.}}%
\label{fig:in_vivo}
\end{figure}

IVIM estimates provided by LSQ and AGP are shown for one subject in Fig.~\ref{fig:in_vivo}.
The estimates by AGP present more homogeneous appearance. 
LSQ seems less reliable in estimating $D^\ast$, since a large number of pixels indicate values on the optimization box constraints. 
Uncertainty maps inferred by AGP illustrate that the estimation is less reliable in areas with low SNR (e.g., background and highly perfused regions) and with significant dephasing artefacts (brain).
Two fitted signals in the brain and placenta show that our method is less biased by dephased data points and better explains the perfusion effect in the brain at low b-values. 

Repeatability of parameters is defined as the intra-subject test-retest variability:
\begin{align}
\small
    \text{VAR}\%=\frac{100}{N}\sum_{i=1,\dots,N} \frac{|\text{TEST}_i-\text{RETEST}_i|}{|\text{TEST}_i+\text{RETEST}_i|/2}
\end{align}
with the number of subjects $N$, where $\text{TEST}_i$ and $\text{RETEST}_i$ are the mean of the parameters in the manually segmented region of interest during repeated acquisitions of subject $i$.
Repeatability results for the placenta region and the averaged mean parameter values across the subjects are listed in Table~\ref{tab:quant_results}. 
The proposed method increases the intra-subject repeatability of $f$ and $D^\ast$ substantially. 
\begin{table}
\setlength{\tabcolsep}{2pt}
\caption{{In vivo fetal MRI, intra-subject repeatability of IVIM in placenta. 
Smaller variance indicated in bold.}}
\label{tab:quant_results}
\centering
   \begin{tabulary}{\linewidth}{l|ccc|ccc}
   \hline
      & \multicolumn{3}{c|}{\bf VAR\%} & \multicolumn{3}{c}{\bf Mean}\\
Method	& $f$ & $D$ & $D^\ast$ &$f$[\%] & $D [\text{mm}^2/\text{s}]$ & $D^\ast [\text{mm}^2/\text{s}]$ \\
\hline
LSQ &24.92  &11.37  &34.35  &28.9 &1.53$\cdot10^{-3}$  &2.72$\cdot10^{-2}$ \\
AGP & \textbf{10.55}  &\textbf{9.67}  &\textbf{12.21}  &30.9 &1.47$\cdot10^{-3}$  &3.57$\cdot10^{-2}$ 
\end{tabulary}
\end{table}

\section{Conclusions}
We propose a novel method (AGP) for probabilistic inverse IVIM mapping, which takes into account non-Gaussian acquisition noise and signal attenuation due to motion induced dephasing. 
A data sampling procedure including dephasing artefacts is defined to learn an inverse mapping using neural networks.
Evaluation on simulated anisotropic IVIM data has showed that, compared to the state-of-the-art LSQ method, AGP improves estimation accuracy by $76\%$ in $f$, $43\%$ in $D$, and $76\%$ in $D^\ast$. 
On in vivo fetal MRI, AGP improves rescan reproducibility by $58\%$ in $f$, $15\%$ in $D$, and $64\%$ in $D^\ast$. 
In addition, our proposed method AGP provides estimation uncertainty, which is essential in clinical diagnosis and experimental design.
Rigorous acquisition models together with more expressive posterior mixtures~\cite{bishop1994mixture} could further expand the range of IVIM applications.

\bibliographystyle{IEEEbib}
\bibliography{refs}

\end{document}